\documentclass[runningheads]{llncs}
\usepackage{graphicx}
\usepackage{color}
\usepackage{amsmath}
\usepackage{multirow}
\usepackage{multicol}
\usepackage{mathtools}
\usepackage{mathrsfs}
\usepackage{subfig}
\usepackage{algorithmic}
\usepackage{textcomp}
\usepackage{algorithm}
\usepackage{graphics}
\usepackage{threeparttable}
\usepackage{amsfonts}
\usepackage{makecell}

\usepackage{multirow}
\usepackage{float}

\usepackage{bm}
\usepackage{array}
\usepackage{colortbl}
\usepackage{pifont}
\usepackage{diagbox}
\usepackage{rotating}
\usepackage{booktabs}
\usepackage{overpic}
\usepackage{contour}
\usepackage[misc]{ifsym} 

\usepackage{url}
\usepackage[colorlinks,linkcolor=blue,anchorcolor=blue,citecolor=blue,]{hyperref}
\def\ourmodel{D$^2$GPLand}

\begin{document}

\newcommand{\tabref}[1]{Table \ref{#1}}
\newcommand{\tickYes}{\bullet}
\newcommand{\cmark}{\ding{51}}
\newcommand{\tickNo}{\hspace{1pt}\ding{55}}
\newcommand{\figref}[1]{Fig. \ref{#1}}
\newcommand{\supp}[1]{\textcolor{magenta}{#1}}
\newcommand{\notsure}[1]{{\textcolor{red}{#1}}}
\newcommand{\secref}[1]{Sec. \ref{#1}}
\newcommand{\sArt}{state-of-the-art}

\newcommand{\ie}{{\emph{i.e.}}}
\newcommand{\eg}{{\emph{e.g.}}}
\newcommand{\etal}{{\emph{et al.}}}

%
\title{Depth-Driven Geometric Prompt Learning for Laparoscopic Liver Landmark Detection}
\titlerunning{Depth-Driven Geometric Prompt Learning for Laparoscopic Liver Landmark Detection}
\author{Jialun Pei\thanks{Equal contribution.}\inst{1},
Ruize Cui\inst{\star 2},
Yaoqian Li\inst{1},
Weixin Si\,\inst{3(\textrm{\Letter})}, 
Jing Qin\inst{2}, \\
Pheng-Ann Heng\inst{1}}

\institute{$^{1}$The Chinese University of Hong Kong, Hong Kong, China\\ 
$^{2}$The Hong Kong Polytechnic University, Hong Kong, China\\
$^{3}$Shenzhen Institute of Advanced Technology, CAS, Shenzhen, China\\
\email{wx.si@siat.ac.cn}}
\authorrunning{Jialun Pei et al.}
%
%
\maketitle              
\begin{abstract}
Laparoscopic liver surgery poses a complex intraoperative dynamic environment for surgeons, where remains a significant challenge to distinguish critical or even hidden structures inside the liver.
Liver anatomical landmarks, \eg, ridge and ligament, serve as important markers for 2D-3D alignment, which can significantly enhance the spatial perception of surgeons for precise surgery.
To facilitate the detection of laparoscopic liver landmarks, we collect a novel dataset called \textbf{L3D}, which comprises 1,152 frames with elaborated landmark annotations from surgical videos of 39 patients across two medical sites.
For benchmarking purposes, 12 mainstream detection methods are selected and comprehensively evaluated on L3D.
Further, we propose a depth-driven geometric prompt learning network, namely \textbf{\ourmodel}.
Specifically, we design a Depth-aware Prompt Embedding (DPE) module that is guided by self-supervised prompts and generates semantically relevant geometric information with the benefit of global depth cues extracted from SAM-based features.
Additionally, a Semantic-specific Geometric Augmentation (SGA) scheme is introduced to efficiently merge RGB-D spatial and geometric information through reverse anatomic perception.
The experimental results indicate that~\ourmodel~obtains state-of-the-art performance on L3D, with 63.52\% DICE and 48.68\% IoU scores.
Together with 2D-3D fusion technology, our method can directly provide the surgeon with intuitive guidance information in laparoscopic scenarios. 
Our code and dataset are available at \url{https://github.com/PJLallen/D2GPLand}.

\end{abstract}
\section{Introduction}
\begin{figure}[t!]
\centering
\includegraphics[width=0.88\textwidth]{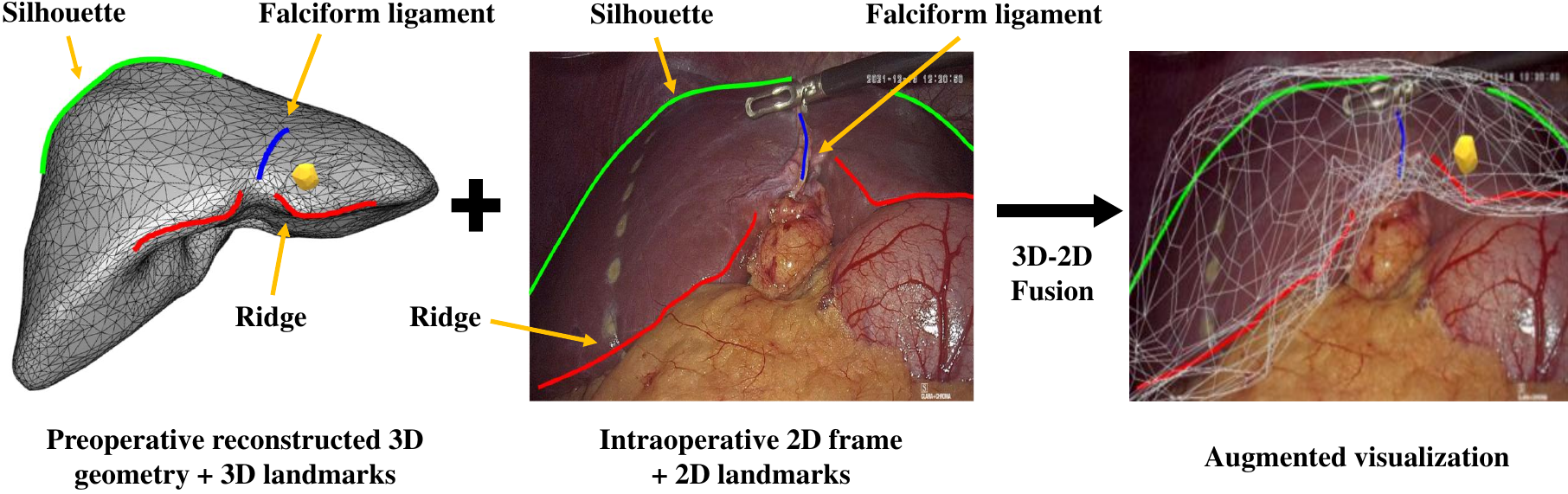}
\caption{Augmented visualization of liver tumor in the laparoscopic video via anatomic landmarks. With consistent anatomical landmarks on 2D frames (middle) and 3D geometry (left), the preoperative 3D anatomy can be overlaid on the intraoperative 2D image for augmented visualization guidance (right).} \label{fig1}
\end{figure}

Laparoscopic liver surgery allows surgeons to perform a variety of less invasive liver procedures through small incisions, enabling faster patient recovery and superior cosmetic outcomes~\cite{surgery}.
However, it is difficult for surgeons to distinguish critical anatomical structures in the complex and variable laparoscopic surgical environment, making it heavily dependent on the experience of the surgeon.
In this regard, augmented reality techniques tailored for laparoscopic liver surgery are urgently desired to provide surgeons with auxiliary information for precise resection and surgical risk reduction. 
The primary step in achieving augmented reality clues is to automatically identify guiding markers on key frames from intraoperative 2D videos and preoperative 3D anatomy samples, respectively, to assist in intraoperative decision-making.
Liver anatomical landmarks, e.g., anterior ridge and falciform ligament, have been validated as effective consistent information for 2D-3D alignment\cite{labrunie2023automatic,labrunie2022automatic}.
As shown in~\figref{fig1}, using 2D and 3D landmarks as references, internal liver structures  are available for intraoperative fusion for enhanced visual guidance.
However, accurate laparoscopic landmark detection remains challenging due to the lack of annotated datasets and how to comprehensively exploit the geometric information in video frames.

Traditionally, landmarks in laparoscopic augmented reality are defined as points or contours~\cite{li2020structured,collins2020augmented}.
In intricate surgical environments, however, the performance of existing structure-based methods suffers from the instability of detection accuracy due to susceptibility to interruptions and tissue deformation together with the lack of global geometric information~\cite{ozgur2018preoperative}.
Additionally, traditional landmarks fail to provide semantic information for precise correspondence between 2D and 3D medical images, which has great importance for estimating cross-dimensional spatial relationships in laparoscopic liver surgery.
To address these challenges, we adapt silhouettes, ridges, and ligaments from laparoscopic video frames as landmarks, which are continuous anatomies with clear semantic features in the preoperative 3D anatomy, facilitating efficient 2D-3D alignment.

\begin{table}[t]
\centering
\renewcommand{\arraystretch}{0.75}
\renewcommand{\tabcolsep}{1.0mm}
\captionof{table}{Statistics of the L3D dataset. \emph{l}, \emph{r}, \emph{s} are ligament, ridge, and silhouette.}
\begin{tabular}{c|c||c|c ||c|c}
\hline
Annotations & Frames & Tumor locations & Cases & Tumor sizes (mm) & Cases \\ 
\hline
$[\emph{l}, \emph{r}, \emph{s}]$ & 1,056 & Quadrate lobe & 7 & 10-19 & 3  \\ 
$[\emph{l}, \emph{r}]$ & 3 & Left lobe & 11 & 20-29 & 10\\ 
$[\emph{r}, \emph{s}]$ & 74 & Right lobe & 21& 30-39 &16 \\
$[\emph{l}, \emph{s}]$ & 19 & Caudate lobe & 0 & 40-49 & 10\\
\hline
\end{tabular}
\label{tab:table1}
\end{table}

However, existing laparoscopic liver landmark datasets lack sufficient annotations for training deep learning-based landmark models~\cite{ali2024objective,rabbani2022methodology,modrzejewski2019vivo}. 
To address the limited sample of liver landmarks, we build the current largest-scale laparoscopic liver landmark dataset, named L3D. 
Specifically, we invite four senior surgeons to select 1152 critical frames from surgical videos of 39 patients at two medical sites, while labeling each frame with three types of semantic landmarks.
Based on the proposed L3D dataset, we contribute a systematic study on 12 mainstream baselines \cite{u-net,labrunie2023automatic,resunet,deeplab,unet++,hrnet,transunet,swinunet,samada,samed,autosam,samlft}.
We observe that existing detection methods concentrate more on semantic feature capture and edge detection while ignoring global geometric features of the liver region, especially the depth information~\cite{koo2022automatic,labrunie2023automatic,labrunie2022automatic}.
Hence, we delve into a straightforward and effective framework that leverages depth maps and pre-trained large vision models to enhance the accuracy of detecting laparoscopic liver landmarks.

In this work, we introduce a depth-driven geometric prompt learning network called~\ourmodel. 
Specifically, we first employ an off-the-shelf depth estimation model to generate depth maps that provide inherent anatomic information. 
Considering that Segment Anything Model (SAM)-based approaches~\cite{kirillov2023segment,wu2023medical} have shown superior performance in extracting global high-level features in surgical scenes, we adopt a pre-trained SAM encoder combined with the CNN encoder to respectively extract RGB multi-level features and depth geometric information. 
Then, a \emph{Bi-modal Feature Unification (BFU)} module is designed to integrate RGB and depth features.
To distinguish highly similar landmark characteristics in laparoscopic liver surgery, we propose a \emph{Depth-aware Prompt Embedding (DPE)} operation to highlight geometric attributes guided by prompt contrastive learning and produce class-aware geometric features.
Moreover, we propose a \emph{Semantic-specific Geometric Augmentation (SGA)} scheme to effectively fuse class-aware geometric features with RGB-D spatial features, where a reverse anatomic attention mechanism is embedded to focus on the perception of anatomical structures and overcome the difficulty of capturing ambiguous landmarks.
Extensive experimental results on the L3D benchmark show that~\ourmodel~achieves a promising performance.
Our method has great potential to be applied in augmented reality-based intra-operative guidance for laparoscopic liver surgery.

\section{L3D Dataset}
To facilitate the detection of laparoscopic liver landmarks, we establish a landmark detection dataset, termed L3D.
Relevant information about patients and annotation is shown in~\tabref{tab:table1}.
To provide enhanced visualization guidance efficiently during the ever-changing surgical environment, we extract key frames from laparoscopic liver surgery videos to annotate liver landmarks according to the suggestions of surgeons.
To this end, four surgeons are invited to select key frames and label them, two of whom perform the labeling and the other two check the labels.
The selection criterion for the keyframes is to allow the surgeon to observe the global view of the liver, which can greatly reduce anatomical misperception during complex laparoscopic liver surgery.
In our dataset, the ridge landmark is defined as the lower anterior ridge of the liver, and the ligament landmark is defined as the junction between the falciform ligament with the liver. In addition, the visible silhouette is also considered as a landmark category.

Our dataset is collected from two medical sites, and all surgeries are liver resections for hepatocellular carcinoma (HCC).
The annotators screen 1,500 initial frames from 39 patient surgery videos with an original resolution of 1920*1080, and retain 1,152 key frames after checking.
We divide all samples in L3D into three sets, where 921 images are used as the training set, 122 images as the validation set, and 109 images as the test set.
To ensure the fairness of the experiment, images from the same patient are not shared across these sets.

\section{Methodology}\label{method}

\begin{figure}[t!]
\includegraphics[width=\textwidth]{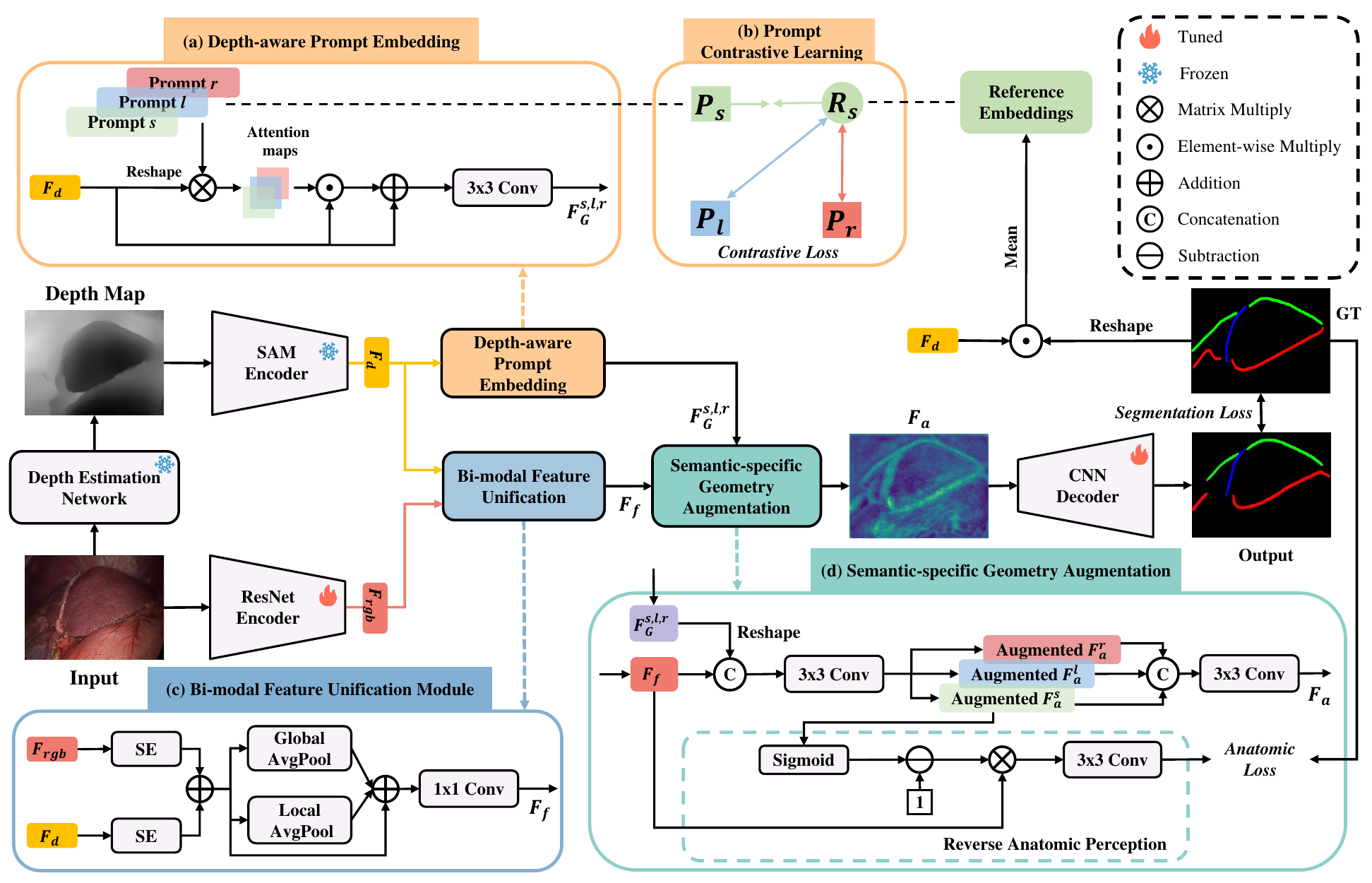}
\caption{Overview of the proposed~\ourmodel. $s,~l$, $r$ denote the three types of landmarks, silhouette, ligament, and ridge, to be detected.} 
\label{overview}
\end{figure}


\figref{overview} outlines the architecture of the proposed~\ourmodel.
%
Our model first takes key frame images from laparoscopic liver surgery as inputs and further generates depth maps using an off-the-shelf depth estimation network (AdelaiDepth~\cite{depth}) as auxiliary inputs to supplement the geometric information. 
%
Then, we employ a ResNet-34 encoder~\cite{resnet} for RGB spatial feature extraction together with a frozen SAM encoder~\cite{kirillov2023segment} for depth geometric cue acquisition.
Notably, the original RGB frames are encoded through a CNN encoder to capture lower-level features for anatomical structure identification, while depth maps mainly provide global shape attributes and geometric insights. 
Thanks to the transformer-based structure and pre-training with large amounts of natural images, the SAM encoder exhibits heightened sensitivity towards global geometric features from the depth modality.
We conduct ablation studies for different encoder combinations in \secref{ablate}.
%
Subsequently, depth feature $F_d$ is passed into the proposed Depth-aware Prompt Embedding (DPE) module to highlight geometric attributes under the guidance of semantic prompts and then output the class-aware geometric features $F_{G}^{s,l,r}$.
%
In parallel, the Bi-modal Feature Unification (BFU) module is applied to incorporate RGB feature $F_{rgb}$ and $F_d$, producing integrated features $F_f$.
%
Then, we interact geometric features $F_{G}^{s,l,r}$ focusing on different landmark categories with the fused RGB-D features through our Semantic-specific Geometric Augmentation (SGA) scheme to obtain augmented unified features $F_a$.
%
Finally, a CNN decoder is used to produce the detection maps. 
%
The following subsections will elaborate on the key components of~\ourmodel.

\subsection{Depth-aware Prompt Embedding}
%
To capitalize on the advantages of pre-trained foundation models while reducing the computational costs for fine-tuning, we maintain the SAM encoder frozen in our model. 
Nonetheless, it still requires further guidance for extracting semantic geometry features related to landmark anatomy.
%
To address this challenge, we propose three randomly initialized efficient class-specific geometric prompts and the DPE module to guide the extraction of geometric information related to different classes from the features derived from the SAM encoder.
As shown in~\figref{overview}(a), we initially execute matrix multiplication between the input $F_d$ and the geometric prompts, generating spatial attention maps to highlight regions associated with specific classes.
%
Moreover, for each attention map, an element-wise multiplication is applied to depth features with a residual operation to obtain class-activated geometric features $F_{G}^{s,l,r}$.
%
%

%
In addition, the proposed DPE module relies on discriminative prompts to guide the class-specific geometry feature extraction.
%
However, it is challenging to learn precise class-specific prompts due to the highly similar landmark characteristics of the liver.
To enhance prompt discriminativeness for better guidance, we apply the contrastive learning technique as illustrated in~\figref{overview}(b).
Here we take the silhouette prompt $P_s$ as an example.
%
Given the ground truth of the silhouette landmark and $F_d$, a dot product is conducted on them, followed by taking the channel-wise mean values to obtain the reference embeddings $R_s$.
Upon obtaining all reference embeddings of the three landmark classes, we modify the NT-Xent Loss~\cite{chen2020simple} as the contrastive loss, formulated as follows:
\begin{equation}
\mathcal{L}_{cl} = {\frac{1}{N}}{\sum\limits_{l\in{L}}}log{\frac{exp(P_l\cdot{R_l}/\tau)}{\sum_{k\in{L}}{exp(P_l\cdot{R_k}/\tau)}}},
\end{equation}
where $N = 3$ is the number of classes, $L = \{s, l, r\}$ denotes the set of all classes, and $\tau$ refers to the temperature-scaled parameter.
%
This contrastive learning strategy enhances the distinctiveness of the class-specific prompt representations.

\subsection{Geometry-enhanced Cross-modal Fusion}

\noindent\textbf{Bi-modal Feature Unification.}
%
To capture holistic landmark features, we propose a BFU module to merge CNN-based lower-level structural features and SAM-based global geometric features.
As depicted in~\figref{overview}(c), we first adaptively adjust the channel weights of $F_{rgb}$ and $F_d$ with Squeeze and Excitation (SE) blocks~\cite{hu2018squeeze} and add them together.
%
Afterward, we embrace the local and global average pooling modules to unify $F_{rgb}$ and $F_d$ at different scales and output the fused feature $F_f$.
%

\noindent\textbf{Semantic-specific Geometry Augmentation.}
%
To further inject the class-activated geometric information from feature $F_{G}^{s,l,r}$ into the fused feature $F_f$, we present the SGA scheme shown in~\figref{overview}(d).
%
%
%
We concatenate each class-specific feature in $F_{G}^{s,l,r}$ with the fused feature $F_f$ respectively, and then obtain the corresponding augmented feature $F_{a}^{s,l,r}$ by 3$\times$3 convolutional block.
%
Subsequently, we concatenate all three semantic geometric features and generate the final augmented feature $F_a$.
%
%
Considering the high similarity between anatomical structure and surrounding tissue features, we also embed a reverse anatomical perception module in the SGA to improve the sensitivity to ambiguous anatomical structures.
%
Inspired by reverse attention\cite{chen2018reverse,pei2022osformer}, we apply a sigmoid function and reverse the attention weights to yield the anatomic attention maps.
%
Afterward, we interplay the attention map with $F_f$ via element-wise multiplication to predict anatomical features.
Here, we use the dice loss as the anatomic Loss $\mathcal{L}_{ana}$ to supervise the anatomic learning.

\subsection{Loss Function}

In addition to the above-mentioned contrast loss and anatomic loss, we also add the segmentation loss $\mathcal{L}_{seg}$ to the overall loss function to supervise the final landmark detection map.
%
In summary, the total loss function can be defined as:
\begin{gather}
\mathcal{L}_{total} = \lambda_{seg}\mathcal{L}_{seg} + \lambda_{cl}\mathcal{L}_{cl} + \lambda_{ana}\mathcal{L}_{ana},~\\
\mathcal{L}_{seg} = \frac{1}{N}\sum_{l\in{L}}(\mathcal{L}_{dice}^{(l)} + \mathcal{L}_{bce}^{(l)}),
\end{gather}
where $\mathcal{L}_{dice}^{(l)}$ denotes the Dice Loss, $\mathcal{L}_{bce}^{(l)}$ denotes the binary cross-entropy (BCE) loss. $\lambda_{seg},~\lambda_{cl},$ and $\lambda_{ana}$ are the balancing parameters for $\mathcal{L}_{seg},~\mathcal{L}_{cl},$ and $\mathcal{L}_{ana}$, respectively.
%
All balancing parameters are set to 1 for optimal performance.

\section{Experiments}

\subsection{Implementation Details}\label{implement}
The proposed \ourmodel~is developed with PyTorch, and the training and testing processes are executed on a single RTX A6000 GPU.
We run 60 epochs for training with a batch size of 4.
A frozen pre-trained SAM-B~\cite{kirillov2023segment} is implemented in the depth encoder.
We resize all the images to 1024$\times$1024 and apply random flip, rotation, and crop for data augmentation.
The Adam optimizer is used with the initial learning rate of 1e-4 and weight decay factor of 3e-5.
In addition, the CosineAnnealingLR scheduler is applied to adjust the learning rate to 1e-6.
For evaluation, we utilize the Intersection over Union (IoU), Dice Score Coefficient (DSC), and Average Symmetric Surface Distance (Assd) as evaluation metrics.


\subsection{Comparison with State-of-the-Art Methods}

\begin{table}[t]
\centering
\renewcommand{\arraystretch}{0.9}
\renewcommand{\tabcolsep}{3.5mm}
\captionof{table}{Comparison with state-of-the-art methods on L3D test set.}
\begin{tabular}{c|l|c|c|c|c}
\hline
  & Models & Model Params  & DSC~$\uparrow$ & IoU~$\uparrow$ & Assd~$\downarrow$ \\ \hline
\multirow{8}{*}{\rotatebox{90}{Non-SAM-based}} & UNet \cite{u-net} & \textbf{7.84M} & 51.39 & 36.35 & 84.94\\ 
                   & COSNet \cite{labrunie2023automatic} &81.24M & 56.24 & 40.98 & 69.22 \\ 
                   & ResUNet \cite{resunet} &69.73M & 55.47 & 40.68 & 70.66\\ 
                   & UNet++ \cite{unet++} & 9.16M & 57.09 & 41.92 & 74.31 \\ 
                   & HRNet \cite{hrnet} &  9.64M  & 58.36 & 43.50 & 70.02  \\ 
                & DeepLabv3+ \cite{deeplab} &43.90M & 59.74 & 44.92 & 60.86\\
                   & TransUNet \cite{transunet} & 71.01M  & 56.81 & 41.44 & 76.16\\
                   & SwinUNet \cite{swinunet} &27.17M & 57.35 & 42.09 & 72.80\\  \hline
\multirow{5}{*}{\rotatebox{90}{SAM-based}} & SAM-Adapter \cite{samada} & 93.93M & 57.57 & 42.88 & 74.31 \\ 
                   & SAMed \cite{samed} &183.55M & 62.03 & 47.17 & 61.55 \\ 
            & SAM-LST \cite{samlft} &183.12M & 60.51 & 45.03 & 68.87 \\                    
                & AutoSAM \cite{autosam} &101.43M & 59.12 & 44.21 &62.49\\ 
 & \textbf{\ourmodel} & 139.03M& \textbf{63.52} & \textbf{48.68} & \textbf{59.38} \\ 
                   \hline
\end{tabular}
\label{tab:results}
\end{table}


\begin{figure}[t!]
\centering
\includegraphics[width=\textwidth]{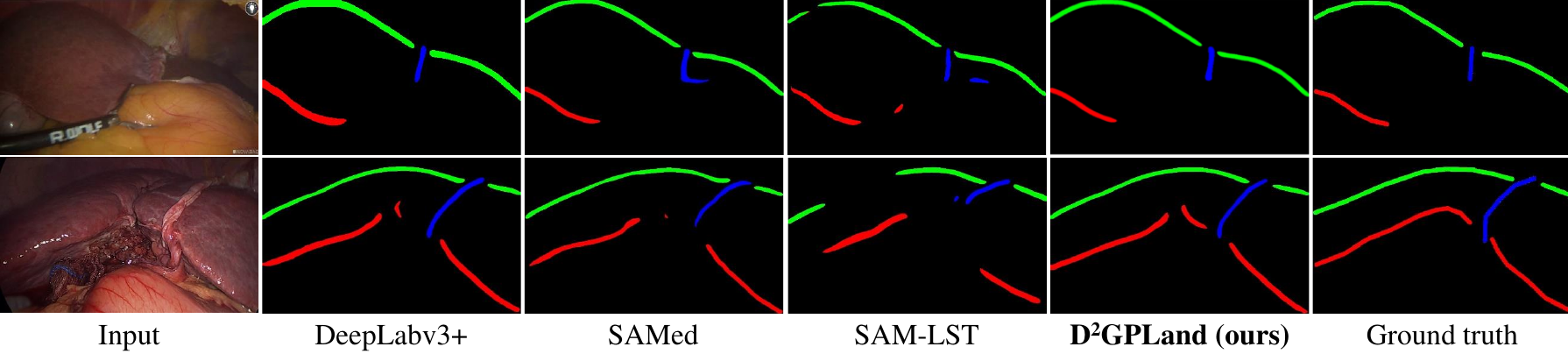}
\caption{Visualizations of our \ourmodel~and competitors on L3D test set.} \label{fig3}
\end{figure}

We compare the proposed \ourmodel~with 12 cutting-edge methods on the L3D test set.
For a fair comparison, these methods are divided into two types: (1) Non-SAM-based models, including UNet~\cite{u-net}, COSNet~\cite{labrunie2023automatic}, ResUNet~\cite{resunet}, DeepLabV3+~\cite{deeplab}, UNet++~\cite{unet++}, HRNet~\cite{hrnet}, TranUNet~\cite{transunet}, and SwinUNet~\cite{swinunet}, and (2) SAM-based models, including SAM-Adapter~\cite{samada}, SAMed~\cite{samed}, AutoSAM~\cite{autosam}, and SAM-LST~\cite{samlft}. 
All compared models were trained to converge with their official implementations.
As shown in Table. \ref{tab:results}, \ourmodel~outperforms competitors on all evaluation metrics. 
Compared to the top-ranked model SAMed, our method improves 1.51\% on DSC, 1.49\% on DSC, and 2.17 pixels on Assd metrics with 44.52M fewer parameters, demonstrating the effectiveness of utilizing depth-aware prompt and semantic-specific geometric augmentation for landmark detection.
Besides, we observe that non-SAM-based methods exhibit inferior performance compared to most SAM-based methods.
It illustrates that the global geometric information extracted by the pre-trained SAM encoder can enhance the perception of landmark features.
%
\figref{fig3} also exhibits the visual results of \ourmodel~and other well-performed methods.
We can see that our method provides more accurate detection of liver landmarks while mitigating the impact of occlusion by other tissues and surgical tools.

\begin{table}[!t]
    \begin{minipage}[t]{0.45\textwidth}
        \centering
        \captionof{table}{Ablations for key Designs.}
        \scalebox{0.9}{
        \begin{tabular}{c|c|c|c|c|c|c}
            \hline
            Methods & BFU & DPE & $\mathcal{L}_{cl}$ & SGA & DSC & IoU\\
            \hline
             M.1 & & & & & 59.34 & 44.90\\ 
             M.2 & $\checkmark$ &  &  & & 61.13 & 46.22\\ 
             M.3 & $\checkmark$ & $\checkmark$ &  &  & 61.98 & 47.12\\
             M.4 & $\checkmark$ & & & $\checkmark$ & 62.20 & 47.20\\
             M.5 & $\checkmark$ & $\checkmark$ & $\checkmark$ & & 62.41 & 47.34\\
             M.6 & & $\checkmark$ & $\checkmark$ & $\checkmark$ & 62.95 & 47.81\\
             Ours & $\checkmark$ & $\checkmark$ & $\checkmark$ & $\checkmark$ & \textbf{63.52} & \textbf{48.68} \\
            \hline
        \end{tabular}
        }
        \label{ablation1}
    \end{minipage}
    \begin{minipage}[t]{0.05\textwidth}
        \begin{tabular}{c}
              
        \end{tabular}
    \end{minipage}
    \begin{minipage}[t]{0.45\textwidth}
        \centering
        \captionof{table}{Ablations for backbones.}
        \scalebox{0.9}{
            \begin{tabular}{c|c|c|c}
            \hline
            Methods & Backbones & DSC & IoU\\
            \hline
            Dual CNN & ResNet-34 & 62.54 & 46.91\\
            Dual SAM & SAM & 62.97 & 47.63\\
            \multirow{2}{*}{SAM+CNN}
            & \multirow{2}{*}{\makecell[c]{ResNet-34(Depth) \\ + SAM(RGB)}} & \multirow{2}{*}{62.83} & \multirow{2}{*}{47.39}\\ 
            & & & \\
           \multirow{2}{*}{CNN+SAM}
            & \multirow{2}{*}{\makecell[c]{ResNet-34(RGB) \\ + SAM(Depth)}} & \multirow{2}{*}{\textbf{63.52}} & \multirow{2}{*}{\textbf{48.68}}\\ 
            & & & \\
            \hline
            \end{tabular}
        }
        \label{ablation2}
    \end{minipage}
\end{table}

\subsection{Ablation Study}\label{ablate}

\noindent\textbf{Ablations for Key Designs.}
Table~\ref{ablation1} shows the contribution of each key design in~\ourmodel~on the L3D test set.
Notably, all variants are trained with the same settings as mentioned in~\secref{implement}.
The baseline (M.1) comprises a ResNet-34 encoder and frozen SAM-B encoder, and we directly concatenate RGB and depth features before feeding them into the decoder.
Overall, each component contributes to the performance of our model in varying degrees.
Specifically, M.2 and M.6 show the effectiveness of our BFU module in merging RGB and depth features.
Based on M.2, M.3 and M.5 sequentially integrate our DPE and contrastive loss $\mathcal{L}_{cl}$ to further enhance the model performance.
Further, M.4 adds the SGA scheme to M.2, resulting in 1.07\% and 0.89\% improvements in DSC and IoU, respectively, indicating the advantages of geometric cues.

\noindent\textbf{Backbone Selections.}
To explore the effect of different backbones in feature extraction across RGB and depth modalities, we conduct additional ablation experiments on L3D with the CNN-based encoder and the SAM-based encoder. 
As shown in~\tabref{ablation2}, \ourmodel~achieves the optimal performance when leveraging the ResNet-34 encoder for RGB inputs and the SAM encoder for depth modality.
This experiment further validates the description in~\secref{method} that the ResNet-34 encoder is more effective in capturing lower-level anatomical structural features while SAM excels in extracting global geometric features.

\section{Conclusion}
This paper proposes a novel geometric prompt learning framework, \ourmodel, for liver landmark detection on key frames of laparoscopic videos.
Our method utilizes depth-aware prompt embeddings and semantic-specific geometric augmentation to explore the intrinsic geometric and spatial information, improving the accuracy of landmark detection.
Moreover, we release a new laparoscopic liver landmark detection dataset, L3D, to advance the landmark detection community.
Experimental results indicate that~\ourmodel~outperforms cutting-edge approaches on L3D, demonstrating the effectiveness of our method in capturing anatomical information in various surgeries.
We hope this work can pave the way for extracting consistent anatomical information from 2D video frames and 3D reconstructed geometries, thereby directly promoting 2D-3D fusion and providing surgeons with intuitive guidance information in laparoscopic scenarios.

\begin{credits}
\subsubsection{\ackname} The work was supported in part by a grant from the Research Grants Council of the Hong Kong Special Administrative Region, China (Project No.: T45-401/22-N), in part by a grant from Hong Kong Innovation and Technology Fund (Project No.: MHP/085/21), in part by a General Research Fund of Hong Kong Research Grants Council (project No.: 15218521), in part by grants from National Natural Science Foundation of China (62372441, U22A2034), in part by Guangdong Basic and Applied Basic Research Foundation (2023A1515030268), in part by Shenzhen Science and Technology Program (Grant No.: RCYX20231211090127030), and in part by Guangzhou Municipal Key R\&D Program (2024B03J0947).

\subsubsection{\discintname}
The authors have no competing interests to declare.
\end{credits}

\bibliographystyle{splncs04}
\bibliography{Paper-0310}
%




\end{document}